\documentclass[journal]{IEEEtran}
\ifCLASSINFOpdf
\else
\fi
\usepackage{amsmath}
\usepackage{booktabs}
\usepackage{multirow}
\usepackage{hyperref}
\usepackage{longtable}
\usepackage{array}
\usepackage{makecell}
\usepackage{threeparttable}
\usepackage{booktabs}
\usepackage{graphicx}
\usepackage{amsfonts}
\usepackage{stfloats}
\usepackage{cite}
\usepackage[justification=centering]{caption} 

\allowdisplaybreaks[4] 
\hypersetup{
	colorlinks = true,
	linkcolor = red,
	anchorcolor = blue,
	citecolor = green
}

\begin{document}
\title
{Multi-view Laplacian Eigenmaps \\
Based on Bag-of-Neighbors \\
For RGBD Human Emotion Recognition}

\author{Shenglan~Liu, \IEEEmembership{Member,~IEEE,}
        ~Shuai~Guo, 
        ~Hong~Qiao, \IEEEmembership{Senior Member,~IEEE}, 
        ~Yang~Wang, ~Bin~Wang, 
        ~Wenbo~Luo, 
        ~Mingming~Zhang, ~Keye~Zhang, and ~Bixuan~Du
\thanks{S. Liu, S. Guo, Y. Wang, B. Wang are with the School of Computer Science and Technology, 
Dalian University of Technology, Dalian 116024, China 
(e-mail: liusl@dlut.edu.cn; guoshuaiabc@mail.dlut.edu.cn; wangyang521@mail.dlut.edu.cn; coding\_rabbit@mail.dlut.edu.cn).}
\thanks{H. Qiao is with the State Key Laboratory of Management and Control for Complex Systems, 
Institute of Automation, Chinese Academy of Sciences, 
Beijing 100190, China (e-mail: hong.qiao@ia.ac.cn).}
\thanks{W. Luo, M. Zhang, K. Zhang, B. Du are with the Research Center for Brain and Cognitive Neuroscience, Liaoning Normal University. 
Dalian 116024, China (e-mail: luowb@lnnu.edu.cn; zmm1001psy@163.com; zhangkeyehaha@126.com; dbx9509@163.com).}
}

\markboth{JOURNAL OF \LaTeX \ CLASS FILES}
{Shell \MakeLowercase{\textit{et al.}}: Bare Demo of IEEEtran.cls for IEEE Journals}

\maketitle

\begin{abstract}
Human emotion recognition is an important direction in the field of biometric and information forensics. 
However, most existing human emotion research are based on the single RGB view. 
In this paper, we introduce a RGBD video-emotion dataset and a RGBD face-emotion dataset for research. 
To our best knowledge, this may be the first RGBD video-emotion dataset. 
We propose a new supervised nonlinear multi-view laplacian eigenmaps (MvLE) approach 
and a multi-hidden-layer out-of-sample network (MHON) for RGB-D human emotion recognition. 
To get better representations of RGB view and depth view, 
MvLE is used to map the training set of both views from original space into the common subspace. 
As RGB view and depth view lie in different spaces, 
a new distance metric bag of neighbors (BON) used in MvLE can get the similar distributions of the two views. 
Finally, MHON is used to get the low-dimensional representations of test data and predict their labels. 
MvLE can deal with the cases that RGB view and depth view have different size of features, even different number of samples and classes. 
And our methods can be easily extended to more than two views. 
The experiment results indicate the effectiveness of our methods over some state-of-art methods. 
\end{abstract}


\IEEEpeerreviewmaketitle

\section{Introduction}

\IEEEPARstart{H}{uman} emotion recognition is an emerging and important area in the field of biometric and information forensics, 
where there has been many significant researches. 
Existing researches on human emotion recognition mainly focus on single view methods, 
such as physiological signals emotion recognition 
\cite{physiological_1,physiological_2,physiological_3}, 
image-based face emotion recognition \cite{facial_emotion_1,facial_emotion_2}, 
speech emotion recognition \cite{speech_emotion_1,speech_emotion_2}, 
and video emotion recognition \cite{video_single_1}. 
For face emotion recognition, both tradition features \cite{facial_emotion_features} 
and deep learning methods \cite{facial_emotion_deep} get great performance. 
And for video-emotion recognition, 
the first step of some researches is frame extraction and face detection 
\cite{multiple_models}, 
so they regard it as another form of face emotion recognition. 
Some other researchers use recurrent neural network \cite{video_single_1} 
or 3D convolution neural network \cite{C3D_1} to recognize human emotion in videos. 
With the popularity of deep learning, neural networks have been widely used in human emotion recognition 
\cite{video_single_1,facial_emotion_1,multiple_models,multiple_features}. 

In many scenes of human biometric recognition, people can be observed at various viewpoints, even by different sensors. 
Recently, multi-view emotion recognition gets more attention, 
combinations of pre-trained models \cite{multiple_models}, features \cite{multiple_features}, 
expressions in face, speech and body \cite{multiple_expressions} 
are also important methods of video-emotion recognition. 

Another fact is that, RGB-D cameras have been widely used in industry and indoor scene. 
RGB view mainly focus on color difference and changes, but depth view mainly focus on spatial information and depth of field. 
Therefore, the combination of RGB view and depth view has great necessity and importance in human emotion recognition. 
In this paper we use RGB-D cameras (Kinect-2.0) to shoot videos and take images of professional human emotion performances, 
then we get an a video-emotion dataset and a image-based face-emotion dataset. 
To our knowledge, most of existing RGB-D human emotion research only focus on image dataset \cite{RGBD_face_1}. 
And compared with the traditional photo vs. sketch dataset used in \cite{PLS,MvDA} and many other researches, 
RGB-D data can be obtained in a large amount easily and contains more information. 
Compared with the popular audio-video based emotion recognition dataset AFEW and image-emotion dataset HAPPEI 
used in Emotion Recognition in the Wild challenge (EmotiW) \cite{EmotiW_2016}, 
our two human emotion datasets are collected under a changeless scene, 
and there is only one actor in a single image or video. 
So we can avoid the influence of environmental disturbance and focus more on human emotion. 

However, human movement or emotion data almost always have some nonlinearity, 
while most of existing multi-view learning methods perform poorly in nonlinear data as they are linear methods. 
As a result, in this paper we propose a new nonlinear method multi-view laplacian eigenmaps (MvLE) to fusion RGB view and depth view, 
as well as improving the recognition performance. 
Multi-view learning, which is also known as data fusion or data integration, 
has three main categories: $(1)$ co-training, $(2)$ multiple kernel learning, and $(3)$ subspace learning. 
MvLE is a subspace learning method based on bag of neighbors (BON) and laplacian eigenmaps (LE) \cite{LE}. 
Assuming that input views are generated from a latent subspace, 
subspace learning is usually used in the task of classification and clustering. 
And the methods of subspace learning can be further grouped into two categories: 
two-view learning methods and multi-view learning methods. 
Besides, methods of each category can run in supervised mode or unsupervised mode, 
depending on whether the category information is used or not. 
Here we introduce some subspace learning methods that are usually used 
in the task of classification \cite{Tao_Survey,Sun_Survey}. 

\textit{Two-view unsupervised methods.} 
Canonical correlation analysis (CCA) \cite{CCA} 
may be the most typical method of subspace learning. 
CCA attempts to find two linear transforms for each view 
such that the cross correlation between two views are maximized. 
In \cite{ICA}, a nonlinear version of CCA was provided. 
Kernel canonical correlation analysis (KCCA) \cite{KCCA} is another improved version of CCA 
which introduces kernel method and regularization technique. 
Fukumizu et al. \cite{KCCA_proof} provided a theoretical justification for KCCA. 
To recognize faces with various poses, partial least squares (PLS) was proposed in \cite{PLS} , 
which can be thought as a balance of projection variance and correlation.

\textit{Two-view supervised methods.} 
Correlation discriminant analysis \cite{CDA} (CDA) is a supervised extension of CCA in correlation measure space, 
which considers the correlation of between-class and within-class samples. 
Inspired by linear discriminant analysis (LDA) \cite{LDA}, 
Tae-Kyun Kim et al. proposed discriminative canonical correlation analysis (DCCA) \cite{DCCA} 
that maximizes the within-class correlations and minimizes the between-class correlations from different views. 
In \cite{MFDA}, Diethe et al. derived a regularized two-view equivalent of fisher discriminant analysis (MFDA) 
by employing the category information. 
In \cite{SVM-2K}, Farquhar et al. proposed a single optimization termed SVM-2K that combines SVM and KCCA. 
Recently, multi-view uncorrelated linear discriminant analysis (MULDA) \cite{MULDA} was proposed 
by combining uncorrelated LDA \cite{ULDA} and DCCA to preserve both the class structures of each view and the correlations between views.

\textit{Multi-view unsupervised methods.} 
Multiview CCA (MCCA) \cite{MCCA} is a multi-view extension of CCA, 
which aims at maximizing the cross correlation of each two views. 
Multiview spectral embedding (MSE) is a multiview spectral-embedding algorithm \cite{MSE}, 
which learns a low-dimensional and sufficiently smooth embedding of all views by preserving the locality in the subspace. 
In \cite{han2012sparse}, Han et al. learned low-dimensional patterns from multiple views using principal component analysis (PCA), 
and proposed a framework of sparse unsupervised subspace learning method.

\textit{Multi-view supervised methods.} 
A multi-view semi-supervised method was proposed in \cite{Multi-view-SVM-2K} to improve the performance of unknown distribution data, 
with a modification for the optimization formulation of SVM. 
In \cite{GMA}, Sharma et al. presented a generic and kernelizable multiview analysis framework (GMA) 
for several known supervised or unsupervised methods. 
But GMA only considers the intra-view discriminant information. 
By reproducing kernel Hilbert space, CCA and PCA, 
Zhu et al. \cite{MKCCA} proposed mixed kernel canonical correlation analysis (MKCCA) that can be implemented in 
multi-view learning and supervised learning. 
Multi-view discriminant analysis (MvDA) \cite{MvDA} 
aims at maximizing the between-class variations and minimizing the within-class variations over all views. 

Recently, many multi-view deep learning methods are proposed for different tasks. 
Hang Su et al. proposed multi-view convolutional neural networks (MVCNN) \cite{MVCNN} for 3D shape recognition, 
which can be regarded as a combination of many CNN networks. 
In \cite{MvDN}, a multi-view deep network (MvDN) was proposed to seek for a non-linear and view-invariant representation 
of multiple views. 

The inherent shortage of two-view methods is that it's not easy to extend them to multi-view problems. 
By using one-versus-one strategy, they have to convert a $n$-view problem to $C_n^2$ two-view problems. 
The main shortage of unsupervised methods is that the label information is not utilized, 
which may limit their performance in the task of classification. 
As mentioned above, preserving the local discriminant structure is an important idea. 
And most of supervised methods above are linear methods 
that aims at optimizing the correlation of classes or views. 
Although some of them can deal with the nonlinear problems by using kernel functions, 
but kernel functions take more calculation, 
and sometimes it's difficult to find a suitable kernel function. 
Compared with traditional methods, Multi-view deep learning are much more time-consuming. 

This paper proposes a multi-view laplacian eigenmaps (MvLE) method based on traditional laplacian eigenmaps (LE) \cite{LE}. 
LE is a nonlinear method that estimates the structure of subspace with the weighted graph $W$. 
We reconstruct the weighted graph $W$ over all all views, 
now the weight of each two samples depends on their bag of neighbors (BON) vectors. 
For each sample of a single view, the $i^{th}$ element of its BON vector means the number of samples 
in this sample's $K$-nearest neighbors which labels are $i$. 
And two samples are ``connected'' if their labels lie on the labels of each other's $K$-nearest neighbors. 
So that the local category discriminant structure can be preserved. 
MvLE learns a common subspace for all views where the ``connected'' points stay as close together as possible. 

Furthermore, MvLE is a supervised method in which category information is used. 
But for the test data, category information is unknown and to be predicted. 
To solve this problem, 
a multi-hidden-layer out-of-sample network (MHON) is proposed based on extreme learning machine (ELM) \cite{ELM}. 
ELM is a feed-forward neural network with a hidden layer, which has extremely fast training speed and high recognition rate. 
After getting the low-dimensional representations of training set with MvLE, 
MHON is trained on the original distributions of training set and their labels, 
with the low-dimensional representations feeding back from the guiding layer. 
By applying robust activation function (RAF) \cite{RAF} in hidden layers, 
the learning capability of MHON is improved. 
Finally, we evaluate MvLE and MHON on the two human emotion datasets mentioned above, 
and show both experimentally and theoretically that our framework 
has a significant improvement compared with some known methods. 
The major contributions of this paper are summarized as follows: 
\begin{enumerate}
\item A new multi-view learning method MvLE is proposed to get the low-dimensional representations of training set. 
\item A new distance metric BON is introduced to get the similar distributions of different views. 
\item A multi-hidden-layer network MHON is proposed to get the low-dimensional representations and predict the labels of test data. 
\item Two new human-emotion RGB-D datasets are collected under psychological principles and methods to evaluate the classification performance of proposed method. 
\end{enumerate}

In the following, Section \ref{related_works} reviews some related works of multi-view learning. 
Section \ref{proposed_method} introduces the proposed methods in detail. 
Section \ref{datasets} introduces the two human-emotion datasets collected by our own. 
The experimental results with qualitative and quantitative evaluations are presented in Section \ref{experiments}, 
followed by a conclusion.



\section{Related Works} \label{related_works}
In this section, we review some excising methods that are related to our works, 
including LDA, LE, CCA, PLS, GMA, MvDA and MvDA-VC. 

\subsection{Notations}
Suppose that we are given the samples from many different views, 
$V^i$ denote the samples of the $i^{th}$ view, 
which locates in $d_i$-dimensional vector space, together with labels $Label^i = [label_1^i, label_2^i, \cdots, label_n^i]$. 
And every $label_k^i \in Label^i$ belongs to the label set $C = \{ 1, 2, \cdots, c \}$. 
The multi-view subspace learning methods aim to find a common subspace for various views. 
Important parameters used in this paper are defined in Table \ref{Para}. 

\begin{table}[!htbp]
\centering
\caption{Definitions of Important Parameters}
\begin{tabular}{p{2cm}p{6cm}}
\toprule
Notation & \hspace{1.5cm} Description \\
\midrule
$V^i\in{\mathbb{R}^{{d_i}\times {n_i}}}$ & all $n_i$ samples of $i^{th}$ view \\
$V_k^i$ & $k^{th}$ sample of $i^{th}$ view \\
$V^{ij}$ & samples of $j^{th}$ class in $i^{th}$ view \\
$V_k^{ij}$ & $k^{th}$ sample of $j^{th}$ class in $i^{th}$ view in the subspace \\
$v$ & the number of views \\

$d_i$ & dimension of samples in $V^i$ \\
$n$ & the number of samples of all views \\
$n_i$ & the number of samples in $V^i$ \\
$n_j$ & the number of samples of $j^{th}$ class of all views \\
$n_{ij}$ & the number of samples of $j^{th}$ class in $V^i$ \\
$c$ & the number of class over all views \\
$W$ & the weight graph of LE \\
$W_i\in{\mathbb{R}^{d_i\times dim}}$ & linear transform of the $V^i$ \\
$w_i\in{\mathbb{R}^{d_i}}$ & basic vector of $W_i$ \\

$Label^i$ & labels of all samples in $V^i$ \\
$label_k^i$ & label of $V_k^i$ \\
$dim$ & dimension of the common subspace \\

$Y^i\in{\mathbb{R}^{dim\times n_i}}$ & samples of $i^{th}$ view in the subspace \\
$y_i\in{\mathbb{R}^{dim}}$ & basic vector of $Y_i$ \\
$Y_k^i$ & $j^{th}$ sample of $i^{th}$ view in the subspace \\

$I$ & the identity matrix \\
$tr(X)$ & the trace of symmetric matrix $X$ \\
\bottomrule
\end{tabular}
\label{Para}
\end{table}

\subsection{Linear Discriminant Analysis}
LDA \cite{LDA} is a linear supervised feature extraction and dimensionality reduction (DR) method of single-view learning. 
It seeks for a linear transform to map the samples from original space to a low-dimension subspace, 
such that the between-class variance is maximized and within-class variance is minimized. 
Let's take $V^i$ as an example: 

\begin{align}
\label{LDA}
\mathop{\max\limits_{w}}\frac{w^TS_bw}{w^TS_ww}
\end{align}

In Eq. \ref{LDA}, $S_b$ and $S_w$ denote the between-class variance and within-class variance, which are calculated as below: 

\begin{align}
\begin{split}
S_w = \sum_{j=1}^c\sum_{k=1}^{n_{ij}}&(V_k^{ij} - \mu_j)(V_k^{ij} - \mu_j)^T \\
S_b = \sum_{j=1}^c n_j(&\mu_j - \mu)(\mu_j - \mu)^T \\
\end{split}
\end{align}

where $\mu_j$ denotes the mean of samples in $j^{th}$ class, 
and $\mu$ denotes the mean of all samples in $V^i$. 
There are many multi-view learning methods extended from LDA, such as ULDA, MULDA and MvDA.

\subsection{Laplacian Eigenmaps}
LE \cite{LE} is one of few nonlinear single-view feature extraction and dimensionality reduction methods. 
Given the $n_i$ samples of $V^i$, LE constructs a weighted graph $W$ to connect the neighboring samples: 

\begin{align}
\begin{split}
W_{ab} = 
\begin{cases}
\emph{exp}(-\frac{\|V_a^i-V_b^i\|_2^2}{t}),&if \ \|V_a^i-V_b^i\|_2^2 < \epsilon \\
0,&else
\end{cases}
\end{split}
\end{align}

With the weighted graph $W$, LE aims at preserving the local information. 
Let $y_j$ denotes the low-dimensional representations of $j^{th}$ samples, 
to choose a good map, the criterion for LE is to minimize the following equation: 

\begin{align}
\label{LE}
\begin{split}
\mathop{\min\limits_{Y}}\frac{1}{2}\sum_{a,b}\|y_a - y_b\|_2^2W_{ij}& = \mathop{\min}tr(Y^TLY) \\
s.t. \  Y^TDY &= I \\
D_{kk} = \sum_{j=1}^{n_i}{W_{jk}},&\ L=D-W
\end{split}
\end{align}

In Eq. \ref{LE}, $D$ is a diagonal matrix, and $L$ is the laplacian matrix. 
This equation can be solve with lagrange multiplier method and eigenvalue decomposition. 
LE can obtain the global optima by building a graph incorporating neighborhood information of the view.

\subsection{Canonical Correlation Analysis}
CCA \cite{CCA} is a typical unsupervised two-view subspace learning methods, with normalization as the first step. 
To get a great low-dimensional common subspace, CCA is usually followed with procedure of dimension-reduction algorithm, 
such as LDA. 
CCA aims to find two transforms $w_1$, $w_2$ for $V^1$ and $V^2$
to project the samples of each view into the common subspace, 
by maximizing the correlation of the two views in the subspace: 

\begin{align}
\begin{split}
\mathop{\max\limits_{w_1,w_2}}w_1^T & V^1V^{2T}w_2 \\
s.t. \  w_1^TV^1V^{1T}w_1 = &1, w_2^TV^2V^{2T}w_2 = 1 \\
\end{split}
\end{align}

With lagrange multiplier method, we can get $w_1$ and $w_2$ by resorting to the eigenvalue decomposition. 
As an unsupervised method, CCA can be regarded as the two-view extension of PCA \cite{PCA} . 
The main limitation of CCA is that $V^1$ and $V^2$ must have the same number of samples. 
In addition, CCA can only deal with the two-view learning case.

\subsection{Partial Least Squares}
PLS \cite{PLS} is an unsupervised two-view subspace learning method, which models $V^1$ and $V^2$ such that: 

\begin{gather}
\begin{split}
\label{PLS}
V^1 = P^TY^1 + E \\
V^2 = Q^TY^2 + F \\
Y^2 = DY^1 + H 
\end{split}
\end{gather}

In Eq. \ref{PLS}, $P\in{\mathbb{R}^{dim\times d_1}}$, $Q\in{\mathbb{R}^{dim\times d_2}}$ are the matrices of loadings. 
$E\in{\mathbb{R}^{d_1\times n_1}}$, $F\in{\mathbb{R}^{d_2\times n_2}}$ 
and $H\in{\mathbb{R}^{dim\times n_2}}$ are the residual matrices. 
Besides, $D\in{\mathbb{R}^{dim\times dim}}$ relates the latent scores of $V^1$ and $V^2$. 
PLS correlates the latent score as well as variation presents of $V^1$ and $V^2$: 

\begin{align}
\begin{split}
\mathop{\max\limits_{w_1,w_2}}y^1y^{2T} = &\mathop{\max\limits_{w_1,w_2}}w_1^T V^1V^{2T}w_2 \\
s.t. \  w_1^Tw_1 = &1, w_2^Tw_2 = 1 \\
V^1 = \ &P^TY^1 + E \\
V^2 = \ &Q^TY^2 + F \\
\end{split}
\end{align}

In a word, PLS tries to correlate the latent score of $V^1$ and $V^2$ as well as capturing the variations of $Y^1$ and $Y^2$, 
while CCA only correlates the latent score. 
And PLS can be solved with iterative method. 
Compared with CCA, PLS is a balance between projection variance and correlation.

\subsection{Generalized Multiview Analysis}
GMA \cite{GMA} is a generic and kernelizable multi-view extension of several supervised or unsupervised two-view learning methods, 
including CCA, PLS, LDA and so on. 
GMA follows the form of quadratically constrained quadratic program (QCQP), 
and chooses to maximize covariance between exemplars of different views as bellow: 

\begin{align}
\begin{split}
\label{GMA}
\mathop{\max\limits_{w_1,\cdots,w_v}}\sum_{i=1}^v \mu_i w_i^TA_iw_i &+ \sum_{i<j} 2\alpha_{ij}w_i^TZ_iZ_j^Tw_j \\
s.t. \  \sum_{i=1}^n \gamma_i &w_i^TB_iw_i = 1 \\
\end{split}
\end{align}

In Eq. \ref{GMA}, $n$ is the number of views, $\mu_i$, $\alpha_{ij}$, $\gamma_i$ are balance parameters, 
and $Z_i$ is the exemplar. 
For different methods, $A_i$, $B_i$ and $Z_i$ have different expressions: 
\begin{itemize}
\item CCA: $A_i = 0$, $B_i = V^iW_iV^{iT}$ and $Z_i = V^i$. 
\item PLS: $A_i = 0$, $B_i = I$ and $Z_i = V^i$. 
\item LDA: $A_i = V^iW_iV^{iT}$, $B_i = V^i(I-W_i)V^{iT}$, $Z_i = M^i$. 
\end{itemize}

For CCA, $W_i = I/n_i$. 
For LDA, $W_i^{kl} = 1/n_{ij}$ if both of $V_k^i$ and $V_l^i$ belong to class $t$, $0$ otherwise, 
and $n_{ij}$ is the number of samples for class $j$ in view $i$. 
$M_i$ defines as the matrix with columns that are class means. 
However, GMA has too many parameters like $\mu_i$, $\alpha_{ij}$ and $\gamma_i$, 
which make it difficult to get a satisfactory model. 
Besides, GMA only considers the intra-view discriminant information.

\subsection{Multi-view Discriminant Analysis}
MvDA \cite{MvDA} is a supervised multi-view learning methods, 
which aims at maximizing the between-class variations and minimizing the within-class variations 
for both intra-view and inter-view samples in the subspace: 

\begin{align}
\begin{split}
\label{MvDA}
\mathop{\max\limits_{w_1,\cdots,w_v}}&\dfrac{tr(S_B^y)}{tr(S_W^y)} \\
S_W^y = \sum_{i=1}^v\sum_{j=1}^c\sum_{k=1}^{n_{ij}}&(Y_k^{ij} - \mu_j)(Y_k^{ij} - \mu_j)^T \\
S_B^y = \sum_{i=1}^c n_i(&\mu_i - \mu)(\mu_i - \mu)^T \\
\end{split}
\end{align}

In Eq. \ref{MvDA}, $\mu$ is the mean of all samples in the subspace, 
and $\mu_j$ is the mean of samples of $j^{th}$ class over all views. 

Consider that different views corresponding to the same objects should have similar structures, 
MvDA with View-Consistency (MvDA-VC) assumes that $V^i\beta_i$ = $V^j\beta_j$, and gets the following optimization equation: 

\begin{align}
\begin{split}
\mathop{\max\limits_{w_1,\cdots,w_v}}&\dfrac{tr(S_B^y)}{tr(S_W^y) + \lambda\sum_{i,j=1}^{v}\|\beta_i - \beta_j\|_2^2} \\
\end{split}
\end{align}

MvDA and MvDA-VC can deal with the cases that different views have different number of samples or classes. 
In addition, MvDA has no parameter to tune, and MvDA-VC only has a balance parameter $\lambda$, 
which makes MvDA and MvDA-VC easier to use in practice. 
But the inherent disadvantage of MvDA and MvDA-VC is that they do not take the inter-view or intra-view information into consideration. 
And for MvDA-VC, the assumption that different views have linear relationship does not always stand.

\section{Proposed Method} \label{proposed_method}

\subsection{Overview}
Inspired by the effectiveness of building global optima and preserving local neighborhoods, 
in this section, we present multi-view laplacian eigenmaps (MvLE) and multi-hidden-layer out-of-sample network (MHON). 
We introduce the basic idea and formulation of MvLE. 
As the RGB view and depth view lie in completely different spaces, 
in MvLE we introduce a new distance metrics called bag of neighbors (BON) to get the similar distributions of the two views. 
BON is based on the label information of $K$-nearest neighbors, 
so that the between-class and within-class discriminant information is included. 
And MvLE can map the training set of both views from original space into common subspace or latent space, 

As the label information of test data is unknown and to be predicted, 
inspired by the work in \cite{QLLP}, a multi-hidden-layer out-of-sample network (MHON) 
is proposed based on ELM \cite{ELM} to solve the problem of out-of-sample extension and predict the labels of test data. 
MHON is trained on the training set of RGB-D views and their labels, 
the input of MHON is the original distributions of RGB-D views, the output of MHON is their labels. 
In the guiding layer of MHON, the low-dimensional representations of training set got by MvLE 
is used as the leading information and feed forward. 
For the test data of RGB-D views, MHON can predict their labels, 
and the low-dimensional representations of test data can be obtained in the guiding layer. 

\begin{figure}[!hbtp]
\begin{center}
\begin{tabular}{c}
\includegraphics[height=6.5cm]{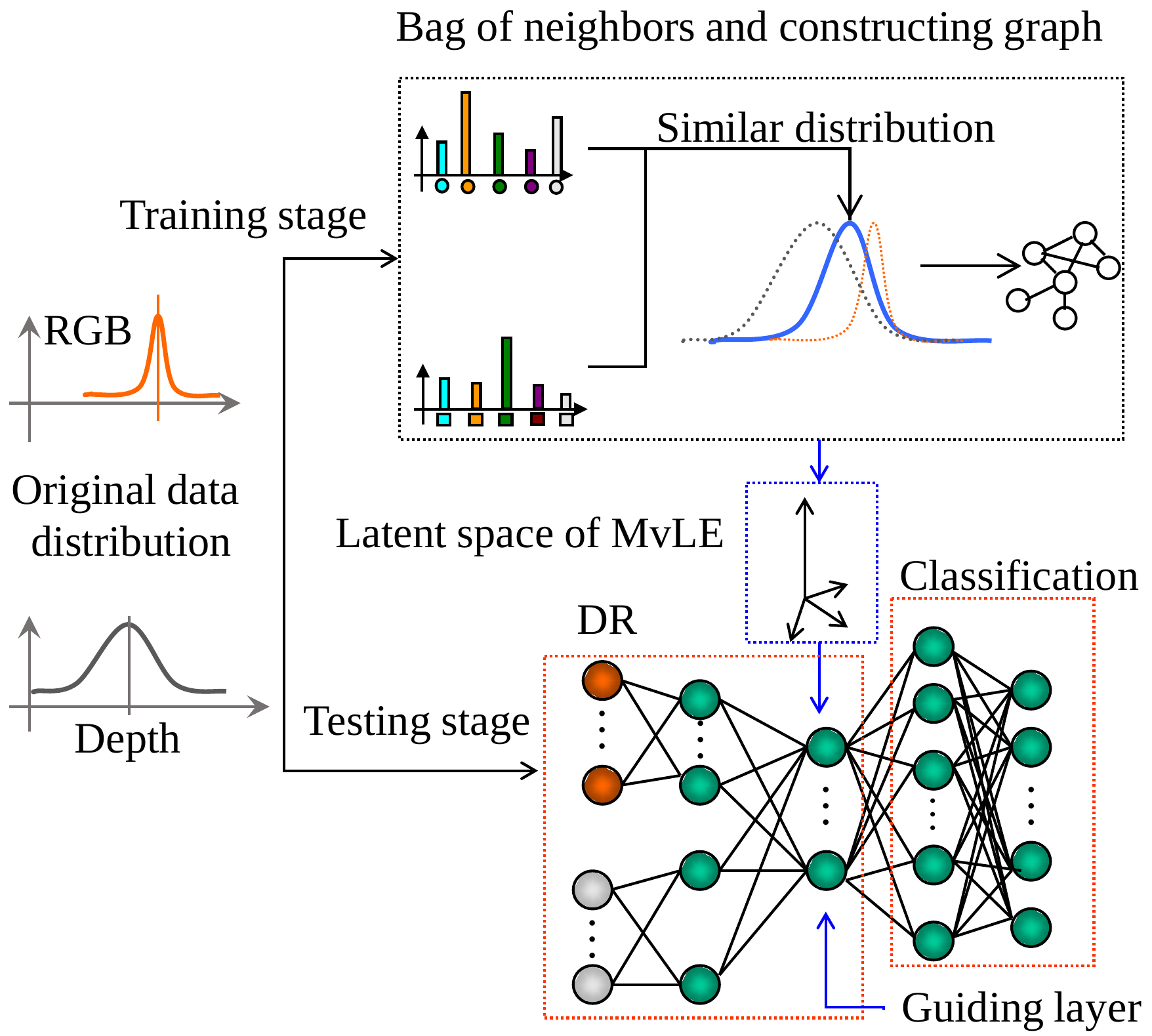}
\end{tabular}
\end{center}
\caption
{ \label{complete_process}
The process of MvLE and MHON. 
}
\end{figure}

The process of our methods is shown in Fig. \ref{complete_process}. 

\subsection{MvLE based on Bag of Neighbors}
For the classification problem, we take two-view learning as an example. 
Suppose that matrices $V^1\in{\mathbb{R}^{{d_1}\times {n_1}}}$ and $V^2\in{\mathbb{R}^{{d_2}\times {n_2}}}$ denote the features of RGB view and depth view, 
$c$ is the number of classes of both views, $n_1$, $n_2$ is the size of training set of each view. 
Note that $n_1$, $d_1$ are not necessarily equal to $n_2$, $d_2$. 

Data normalization is the first step. 
And then, each sample is represented with a bag of neighbors (BON) vector which has a length of $c$. 
Let $BON_k^i = [x_1, x_2, \cdots, x_c]$ denotes the BON vector of sample $V_k^i$, 
where $x_t$ denotes the number of samples which are labeled as class $t$ in the $K$-nearest neighbors of sample $a$. 
The $K$-nearest neighbors depend on Euclidean distance. 
In terms of $V_a^i$ and $V_b^i$, the Euclidean distance is defined as follow: 

\begin{gather}
distance_{ab}^i = \|V_a^i - V_b^i\|_2
\end{gather}

\begin{figure}[!hbtp]
\begin{center}
\begin{tabular}{c}
\includegraphics[height=5cm]{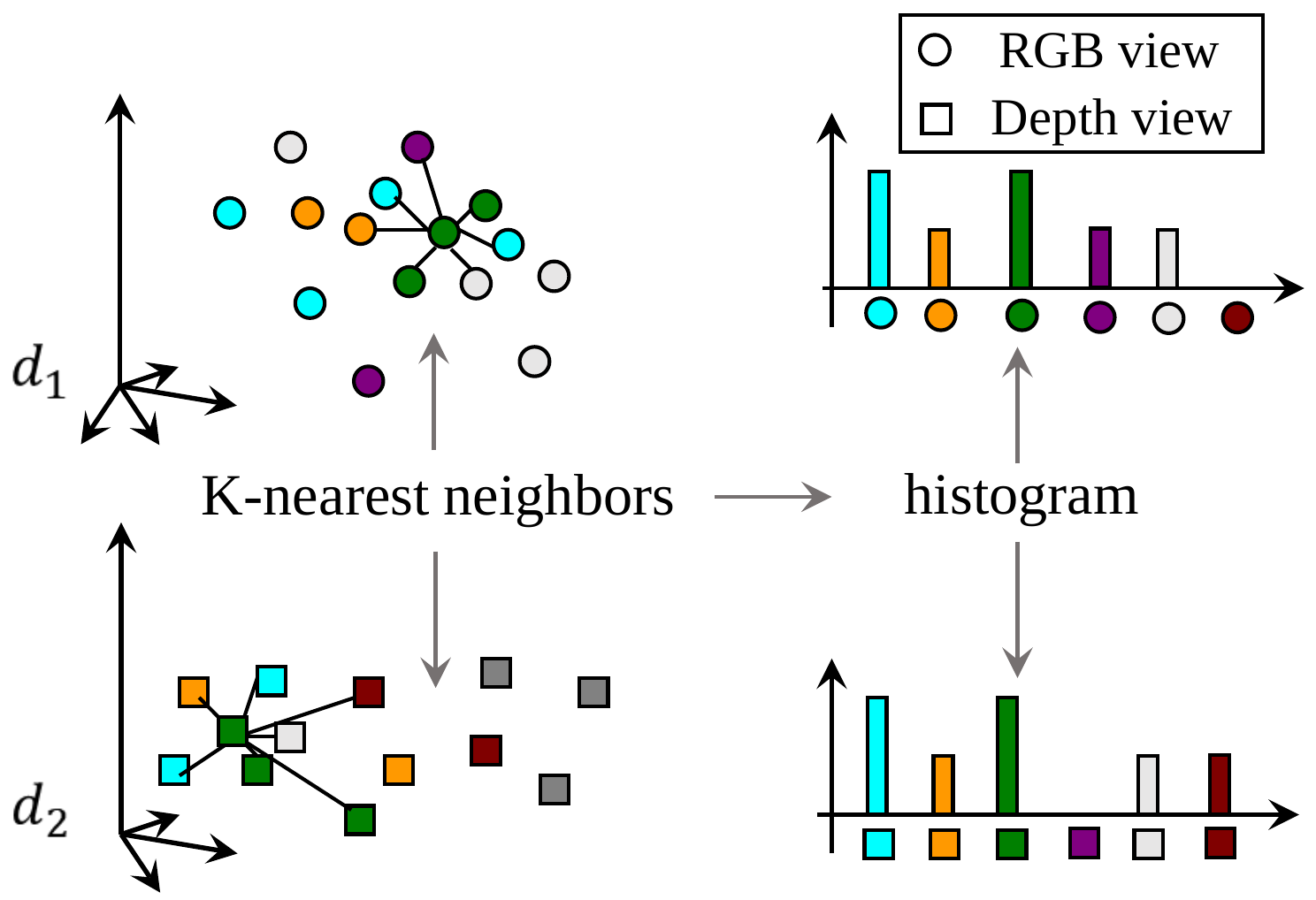}
\end{tabular}
\end{center}
\caption
{ \label{BON}
Bag of neighbors for example. 
}
\end{figure}

By introducing BON, samples of each view can get similar distributions. 
Furthermore, let $W\in{\mathbb{R}^{(n_1+n_2)\times (n_1+n_2)}}$ denotes the new weight matrix of the proposed method. 
Actually, $W$ can be divided into four parts: $\left[W^{11}, W^{12}; W^{21}, W^{22}\right]$, 
indicating four inter-view or intra-view similarity measures. 
And dimensions of them are $n_1\times n_1$, $n_1\times n_2$, $n_2\times n_1$, $n_2\times n_2$ respectively. 
That is: 

\begin{gather}
W = {
\left[ \begin{array}{cc}
W^{11} & W^{12} \\
W^{21} & W^{22} \\
\end{array} 
\right ]}
\xrightarrow {size} {
\left[ \begin{array}{cc}
n_1\times n_1 & n_1\times n_2 \\
n_2\times n_1 & n_2\times n_2 \\
\end{array}
\right ]}
\end{gather}

Compared with traditional laplacian eigenmaps, 
the weighted matrix $W$ of the proposed method depends on $BON$ vectors got above. 
Moreover, four parts of $W$ are calculated respectively. 
Let $Leighb_k^i$ denotes the labels of $K$-nearest neighbors of $V_k^i$, and $label_k^i$ denotes the label of $V_k^i$. 
For the $(a, b)$ element of $W^{ij}$ above, here $a,b \in \{1, 2\}$. 
If $label_a^i \in Lneighb_b^j$ and $label_b^j \in Lneighb_a^i$, 
we would think sample $V_a^i$ is ``connected'' with sample $V_b^j$, no matter they are inter-view samples or intra-view samples. 
Then, BON is naturally used to measure this weight: 

\begin{gather}
W_{ab}^{ij} = \emph{exp}(-\dfrac{\|BON_a^i - BON_b^j\|_2^2}{t})
\end{gather}

In this formula, $t$ is an adjustable constant, 
which we set as $c$ in the follow-up experiments. 
And if $label_a^i \notin Lneighb_b^j$ or $label_b^j \notin Lneighb_a^i$, they are not ``connected'': 

\begin{gather}
W_{ab}^{ij} = 0
\end{gather}

The new distance metric BON can not only overcome the difference between views, 
but also introduce category discriminant information. 
Compared with LDA-based methods that aim at maximizing between-class variance and minimizing within-class variance, 
such as MvDA and MULDA, MvLE tries to minimize the distance between samples that are ``connected''. 
In the proposed method, samples of different classes are almost impossible to be marked as ``connected''. 
Accordingly, BON is more insensitive to outliers and noise. 

After getting $W$, subsequent steps are similar to traditional laplacian eigenmaps. 
Suppose that $Y\in{\mathbb{R}^{(n_1+n_2)\times dim}}$ denotes the features after fusion, 
the first $n_1$ vectors of $Y$ is the low-dimensional representations of $V^1$, 
and the last $n_2$ vectors of $Y$ is the low-dimensional representations of $V^2$. 
Let $y_a$ and $y_b$ denote two vectors of $Y$, 
$v_a$ and $v_b$ denote the original distribution that corresponding to $y_a$ and $y_b$. 
We try to ensure that if $v_a$ and $v_b$ are ``connected'' and the weight between them is low, 
$y_a$ and $y_b$ should stay close as well. 
So the objective function can be written as follow: 

\begin{align}
\begin{split}
\mathop{\min\limits_{Y}}&\xi(Y) \ s.t. \  Y^TDY = I \\
\xi(Y) &= \sum_{a,b}\|y_a - y_b\|_2^2W_{ab} \\
&= \sum_{a,b}(y_a^2 + y_b^2 - 2y_ay_b)W_{ab} \\
&= 2tr(Y^TLY)
\end{split}
\end{align}

The problem boils down to computing eigenvalues and eigenvectors for the generalized eigenvector: 

\begin{gather}
\begin{split}
Ly^i =& \lambda D_{ii}y^i \\
D_{ii} = \sum_{j=1}^{n_1+n_2}{W_{ji}},& \ L=D-W
\end{split}
\label{LyDy}
\end{gather}

In Eq. \ref{LyDy}, $D$ is a diagonal weight matrix, and $L$ is the Laplacian matrix. 
Now let $y^1, y^2, \cdots, y^{n_1+n_2}$ ordered by eigenvalues in ascending order denote the solution of Eq. \ref{LyDy}. 
And then $Y$ is given by $[y^2, y^3, \cdots, y^{dim+1}]$, 
because $y^1$ corresponds to the smallest eigenvalue which value is $0$. 

Let $Y^1$ denotes the first $n_1$ rows of $Y$, $Y^2$ denotes the last $n_2$ rows of $Y$. 
Finally, $Y^1$ is regarded as features of $V_1$ after fusion and dimensionality reduction, 
and $Y^2$ is regarded as features of $V_2$ after fusion and dimensionality reduction. 
In this manner, our method would not be affected by the size of different views. 

If more than two views are given, we just need to build the weight graph like the following matrix. 

\begin{gather}
W = {
\left[ \begin{array}{cccc}
W^{11} & W^{12} & \cdots & W^{1n} \\
W^{21} & W^{22} & \cdots & W^{2n} \\
\vdots  & \vdots  & \vdots & \vdots  \\
W^{n1} & W^{n2} & \cdots & W^{nn} \\
\end{array} 
\right ]}
\end{gather}

And after getting $Y$, the first $n_1$ rows of $Y$ are regarded as $Y^1$, the second $n_2$ rows are regarded as $Y^2$, $\cdots$, 
the last $n_n$ rows are regarded as $Y^n$. 

MvLE builds a global weight graph over all views to incorporate the inter-view and intra-view neighborhood information. 
The size of global graph in this paper is equal to the number of samples of all views. 
With the interaction of different views, samples of each view can get appropriate representations in the subspace, 
which helps to get a better performance in classification. 
As far as we are concerned, there are few researches focusing on building global graph in multi-view learning. 
Most of existing methods like CCA, PLS, and MvDA did not make full use of inter-view and intra-view information. 
Another advantage of global graph is that, 
the multi-view locality-preserving character of MvLE makes it relatively insensitive to outliers and noise. 
But the time complexity and spatial complexity tend to be high as well. 

\subsection{Multi-hidden-layer Out-of-sample Network (MHON)
} 

As a supervised nonlinear multiview learning method, 
category information is used in BOW to measure the weight between samples. 
However, category information is only given for the training dataset, 
for the test dataset, category information is to be predicted. 
Another fact is that, MvLE cannot get a linear transform for each view. 
In \cite{QLLP}, a nonlinear manifold learning framework QLLP was proposed by Shenglan Liu et al. , 
they chose a small subset of original data to learn the explicit mapping function 
from original data to the low-dimensional coordinates. 
Manifold learning assumes that high-dimensional input data lie on a low-dimensional manifold. 
And QLLP preserves the local geometry structure as well as the true manifold structure of original space. 

\begin{figure}[!hbtp]
\begin{center}
\begin{tabular}{c}
\includegraphics[height=4.5cm]{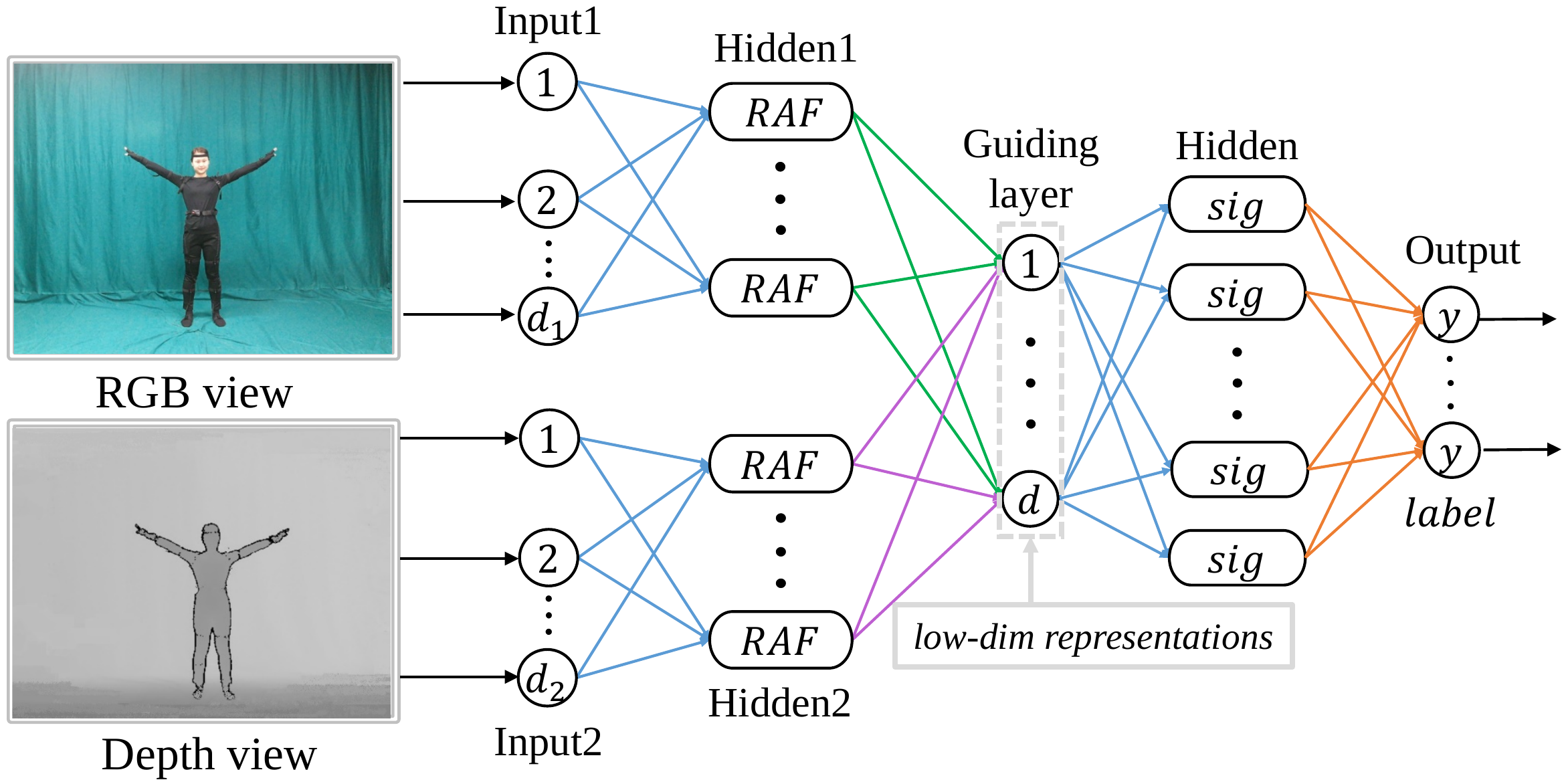}
\end{tabular}
\end{center}
\caption
{ \label{network}
The overview of MHON. 
}
\end{figure}

Inspired by their work, here we propose a multi-hidden-layer out-of-sample network (MHON) 
to get the low-dimension representations of test data and predict their labels, 
as the Fig. \ref{network} shows. 
MHON is trained on the original distributions of RGB-D views and their labels. 
In the guiding layer of MHON, low-dimensional representations of training set 
is used as the leading information and feed forward. 
For the test data, the input of MHON is original distributions of RGB-D views, 
MHON can predict their labels in the last layer, 
and the low-dimensional representations are got in the guiding layer. 

In \cite{RAF}, robust activation function (RAF) is proved to be beneficial to the performance of ELM. 
As a continuous, monotonic and nonlinear active function, 
RAF is used in the first hidden-layer of MHON to improve the recognition performance. 
For the second hidden-layer, we use sigmoid active function to fulfill the task of classification. 


If more than two views are given, the input of MHON is the original distributions of all views. 
Their low-dimensional representations and labels are got in the guiding layer ant output layer respectively. 





\section{Human Emotion Datasets} \label{datasets}

The majority of existing human-emotion datasets suffer from two disadvantage: 
(1) The videos or images in existing datasets could not get rid of the influence of environment. 
(2) The information provided by a single RGB view seem to be deficient. 
In this section, we introduce a new RGB-D video-emotion dataset and a new RGB-D face-emotion dataset 
that are collected at a changeless scene. 
Compared with RGB view that mainly focus on color difference, 
depth view has unique advantages by introducing spatial and depth information of the field. 
The combination of RGB view and depth view would has great necessity and importance to human emotion recognition. 
As far as we are concerned, there are few RGB-D video-emotion datasets in existence. 
AFEW \cite{EmotiW_2016} is a popular video-emotion dataset composed by videos from movies and reality TV shows. 
In contrast to AFEW, the video-emotion dataset is designed under psychological principles and well-designed scripts, 
and there is only one person in an video or image.

\subsection{Video-Emotion Dataset}
The video-emotion dataset consists of over 4k (4 thousand) clips of RGB videos and 4k clips of depth videos that correspond to each other, 
and each video has a length of 6 seconds and a resolution of 702$\times$538. 
It contains the following 7 emotion classes: angry, disgusted, fearful, happy, neutral, sad, and surprised. 
As a whole-body video-emotion dataset, it also has some significance in human-emotion expression from the view point of psychological \cite{why_bodies} . 
The video-emotion dataset is collected under psychological methods and principles, 
firstly, 
we designed a number of 6-seconds length scenes that can show one of the emotions above. 
For example, jumping and dancing with joy means someone is happy, wiping tears and sobbing means someone is sad. 
After that, at least 200 people are asked to grade on these scenes. 
We then know which scenes can show human emotion better. 
At last, we selected 6 highest score scenes for each emotion as the final scripts. 

\begin{figure}[!hbtp]
\begin{center}
\begin{tabular}{c}
\includegraphics[height=5cm]{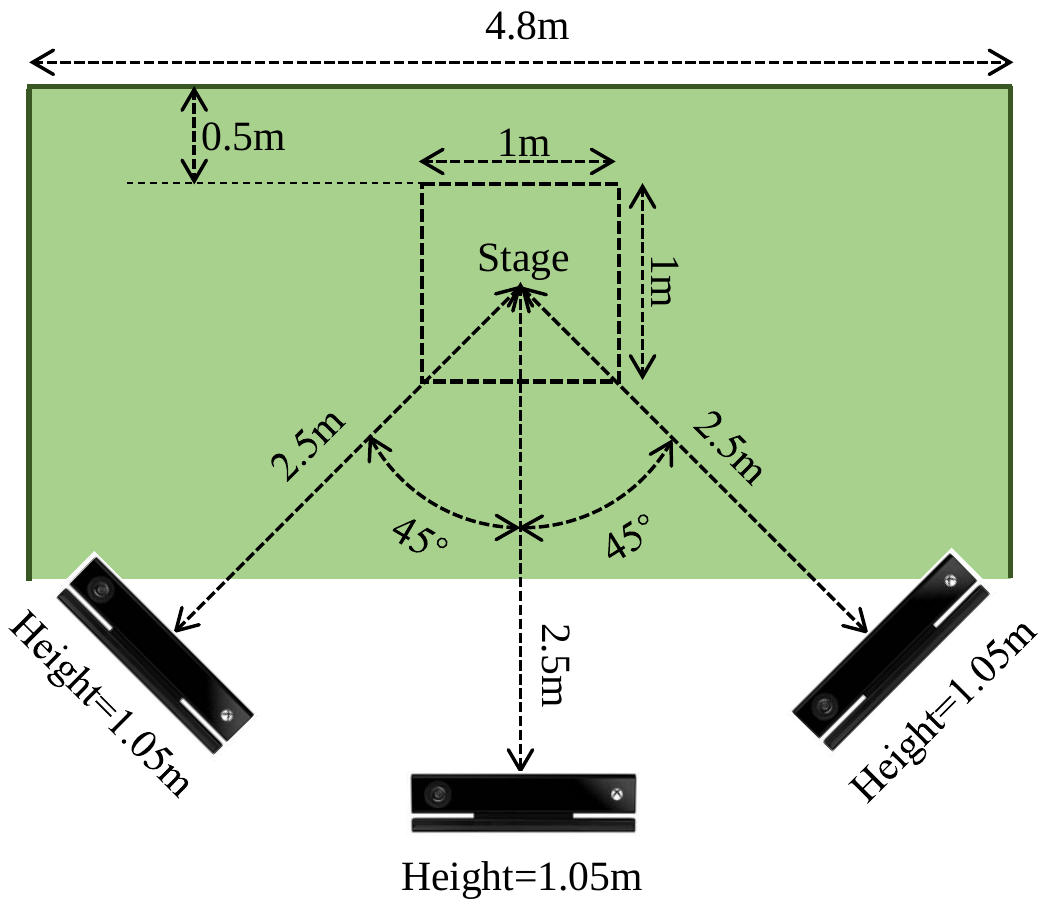}
\end{tabular}
\end{center}
\caption
{ \label{stage}
The scene arrangement of video-emotion dataset. 
}
\end{figure}

\begin{figure}[!hbtp]
\begin{center}
\begin{tabular}{c}
\includegraphics[height=5.6cm]{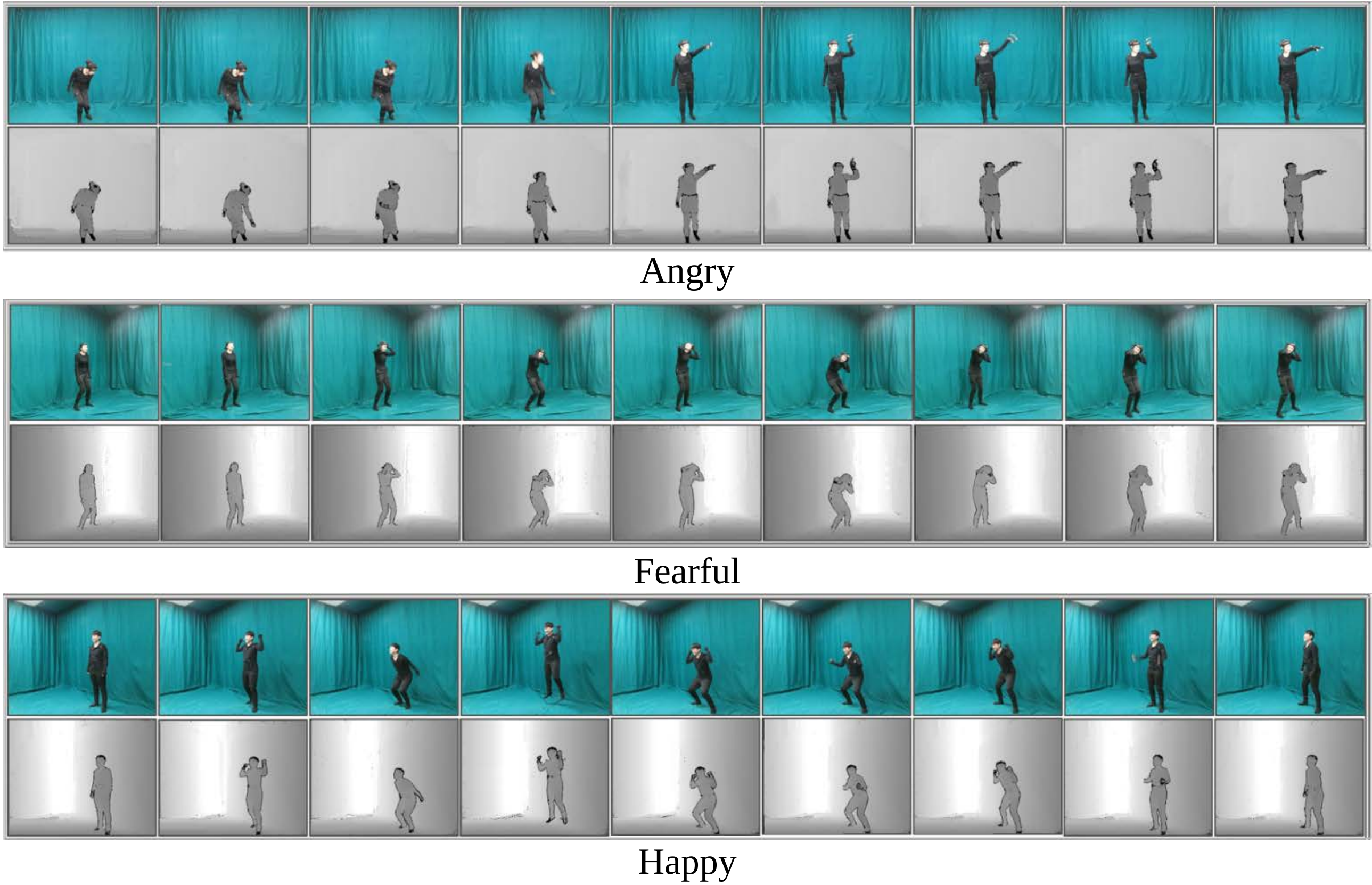}
\end{tabular}
\end{center}
\caption
{ \label{videodataset}
Three examples in video-emotion RGB-D dataset. 
}
\end{figure}

Furthermore, we employed 24 professional actors to perform these scripts. 
The background color of the scene is green, 
and actors perform the scripts at a $1$ square meter stage which is centered at the scene. 
To record their performances, we have 3 Kinect-2.0 cameras shooting RGB-D videos at the same time, 
which are placed at front, left, and right of the stage, as Fig. \ref{stage} shows. 
Actors may perform a script more than one time with different body movements. 
After cutting and editing, we finally get a video-emotion dataset of 7 emotions and 14 hours of RGB-D clips. 
Fig. \ref{videodataset} shows three examples of this dataset, 
and each example has 9 discontinuous frames of a RGB clip and a depth clip that correspond to each other.

\subsection{Face-Emotion Dataset}
The face-emotion dataset includes about 1k RGB face emotion images and 1k depth face emotion images that correspond to each other. 
Less than the video-emotion dataset mentioned above, the face-emotion dataset has 6 emotion classes: 
angry, afraid, happy, neutral, sad, surprised. 
We get 69 volunteers to perform all these emotions with facial expressions from 5 different viewpoints, 
which are front, up, down, left, and right. 

In addition, a Kinect-2.0 camera is used to take RGB-D images of the facial emotion. 
To crop out the background information of the scene, 
we use Kinect-2.0 to detect the position of head and neck of the actor in the image. 
Then we draw a square centered in the position of head in each image, 
and the width of which depends on the distance between head and neck. 
With this square, we crop out the background and get the facial emotion images. 
So that every volunteer have 30 RGB images and 30 depth images taken, 
the resolution of which is about 150$\times$110. 
At last, we get a face emotion dataset of 6 emotion classes, 1k RGB images, and 1k depth images. 

\begin{figure}[!hbtp]
\begin{center}
\begin{tabular}{c}
\includegraphics[height=2.7cm]{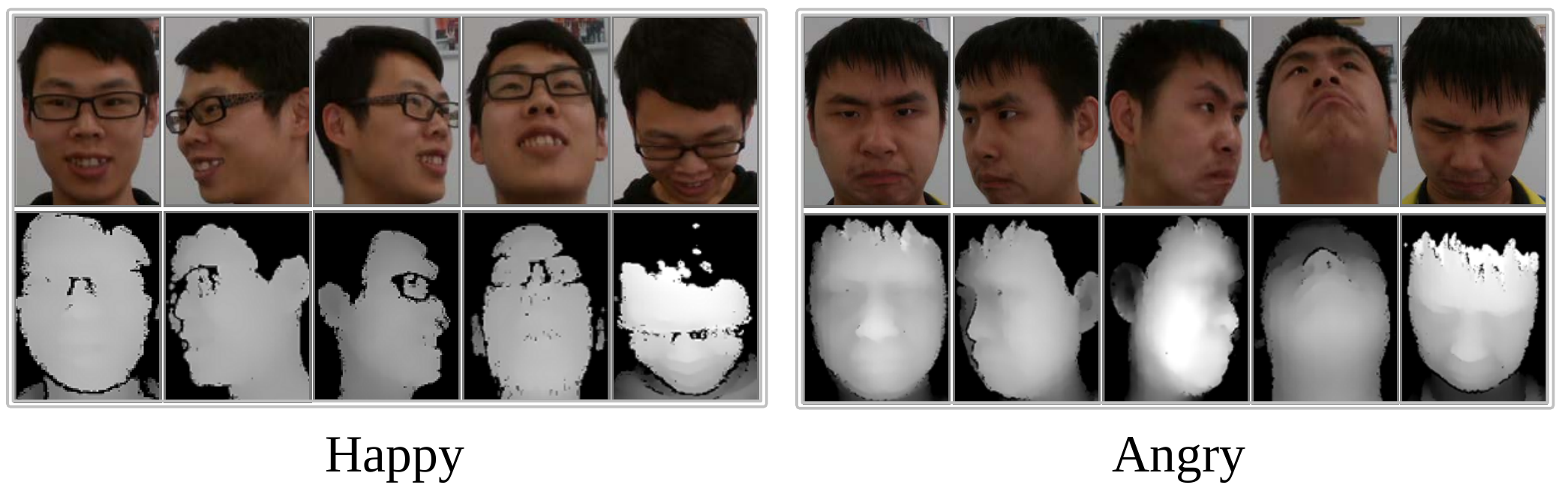}
\end{tabular}
\end{center}
\label{facedataset}
\caption
{ Two examples in face-emotion RGB-D dataset.}
\end{figure}

\section{Experiments} \label{experiments}

In this section, we evaluate the proposed method on the video-emotion dataset and face-emotion dataset introduced above. 
Firstly, we extract C3D features of the video-emotion dataset, and VGG16 \cite{VGG16} features of the face-emotion dataset. 
By introducing intra-class variance $S_W$ and inter-class variance $S_B$, 
we illustrate that both C3D features and VGG16 features are nonlinear features. 
After that, the quantitative comparisons of average accuracy are presented between our new method, CCA-LDA, PLS, GMA, MvDA and MvDA-VC.

\subsection{Features Extraction and nonlinearity analysis}

Taking videos as sequences of frames, 3-Dimensional convolutional neural networks (3D-CNN) 
can capture the spatial and temporal dimensions along with discriminative information. 
As a popular and effective method for spatiotemporal feature learning and video analyzing, 
3D-CNN has been widely used in many researches \cite{C3D_1,Facebook_C3D_3} . 
In addition, C3D-1.0 \cite{Facebook_C3D_1,Facebook_C3D_2,Facebook_C3D_3} trained on UCF101 \cite{UFC101} is 
a modified version of BVLC\_caffe to support 3D-CNN. 
We make a fine-tune on C3D-1.0 and extract features of our video-emotion dataset. 
Besides, convolution neural network has been proved to be extreme useful in image classification, 
we use the classical network VGG16 \cite{VGG16} to extract features of our face-emotion image dataset. 

To evaluate the effectiveness and nonlinearity of the features extracted above, 
we calculate average intra-class variance $S_W$ and inter-class variance $S_B$ for RGB-D features of the two datasets, 
which are defined as below: 

\begin{gather}
S_W = \frac{1}{c}\sum_{i=1}^{c}\sum_{x\in{X_i}}{\frac{1}{n_i - 1}\|x - \mu_i\|_2^2} \\
S_B = \frac{1}{n - 1}\sum_{i=1}^{c}{n_i\|\mu_i - \mu\|_2^2} 
\end{gather}

where $\mu$ denotes the mean of all samples, $\mu_i$ denotes mean of samples in class $i$, 
$N_i$ denotes the neighborhood of sample $i$, 
and $\alpha_{ij}$ denotes the angle between $x_j$ and its orthogonal projection. 
$S_W$ and $S_B$ measure how far the within-class and between-class samples spread out of their mean. 
If the value of $S_W$ and $S_B$ is high, within-class samples and between-class samples would be very different. 
Therefore, the nonlinearity tend to be fine with high values of $S_W$, or a low value of $S_B$. 
We calculate $S_W$ and $S_B$ of RGB data and depth data for C3D features and VGG16 features in Table \ref{SWB}, 
with RGB data and depth data individually. 

\begin{table}[!htbp]
\centering
\caption{$S_W$ and $S_B$ of features of two datasets}
\begin{tabular}{p{1cm}<{\centering}p{1cm}<{\centering}p{1cm}<{\centering}p{1cm}<{\centering}p{1cm}<{\centering}}
\toprule
\multirow{2}{*}{Variance} &
\multicolumn{2}{c}{Video-Emotion Dataset} & \multicolumn{2}{c}{Face-Emotion Dataset} \\
\cmidrule(lr){2-3} \cmidrule(lr){4-5}
& RGB & depth & RGB & depth \\
\midrule
$S_W$ & 0.8324 & 0.8948 & 0.1431 & 0.1814 \\
$S_B$ & 0.0093 & 0.0052 & 0.5970 & 0.5257 \\
\bottomrule
\end{tabular}
\label{SWB}
\end{table}

\begin{figure}[!hbtp] 
\begin{center}
\begin{tabular}{c}
\includegraphics[height=6cm]{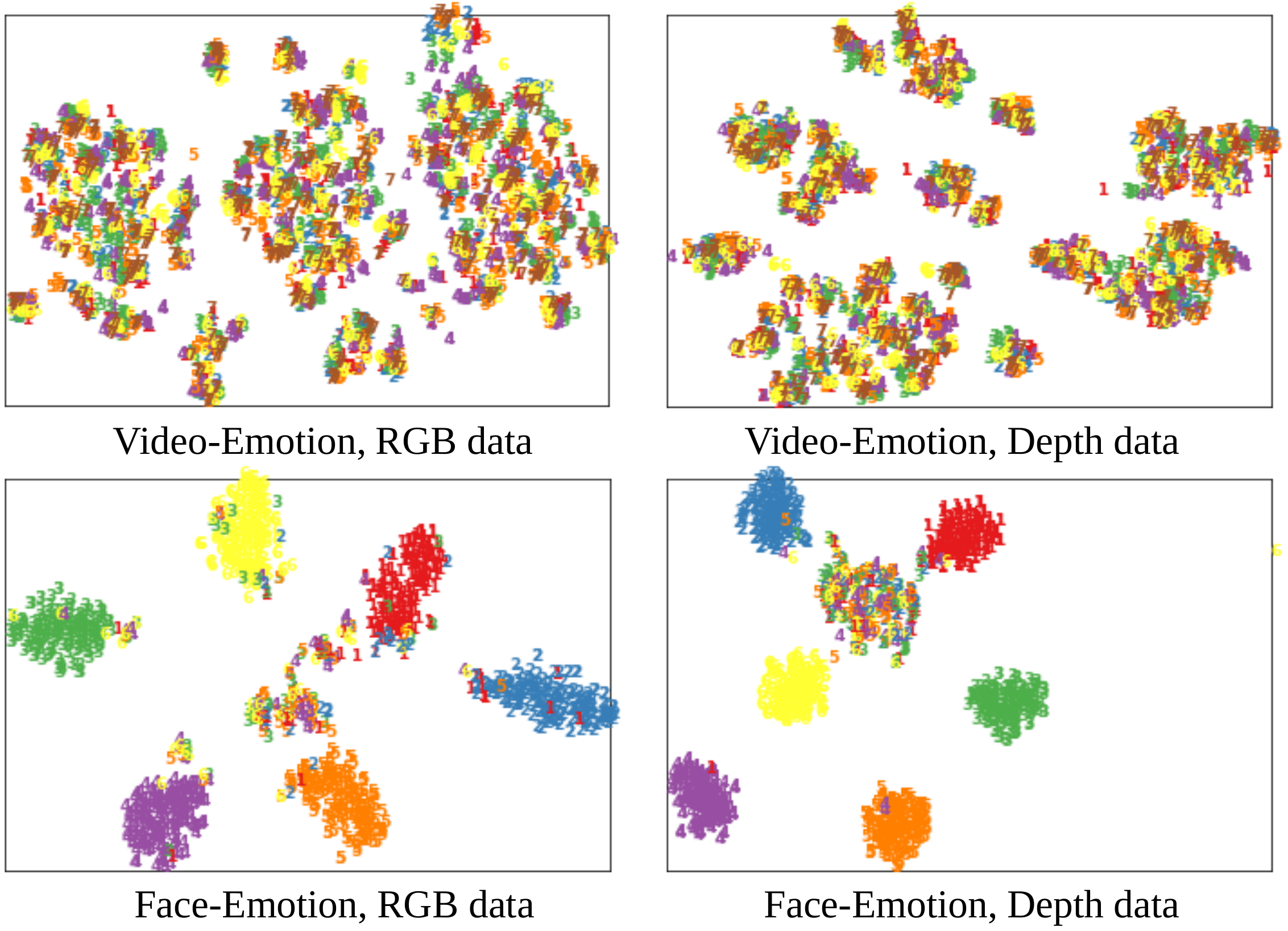}
\end{tabular}
\end{center}

\caption
{ \label{fig:tsne}
Visualizing with t-SNE. 
}
\end{figure}

As seen, C3D features of video-emotion dataset have a high value of $S_W$ and an extremely low value of $S_B$, 
on the contrary of face-emotion dataset. 
This shows that C3D features of video-emotion dataset have great nonlinearity, 
but the nonlinearity of VGG16 features of face-emotion image dataset is not very well. 
Furthermore, C3D and VGG16 features of video-emotion and face-emotion dataset are visualized using t-SNE \cite{tsne} in Fig. \ref{fig:tsne}, 
in which samples of each class are denoted in color-coded figures. 
Fig. \ref{fig:tsne} confirm the analyses of nonlinearity above. 

\subsection{Comparisons And Analyses}
\begin{figure*}[!hbtp] 
\begin{center}
\begin{tabular}{c}
\includegraphics[height=5.2cm]{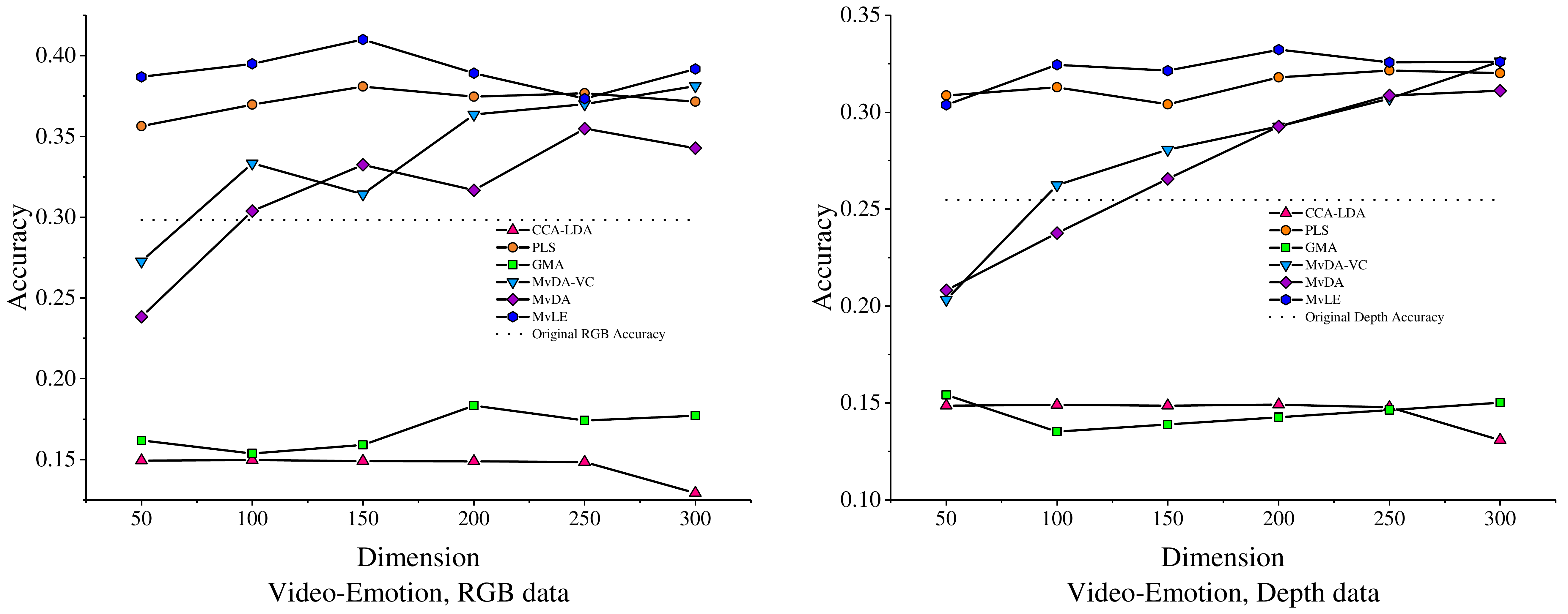}
\end{tabular}
\end{center}
\caption
{ \label{fig:video}
Change curve of comparison methods on different target dimension for the RGB-D video-emotion dataset. 
}
\end{figure*}

\begin{figure*}[!htbp] 
\begin{center}
\begin{tabular}{c}
\includegraphics[height=5.2cm]{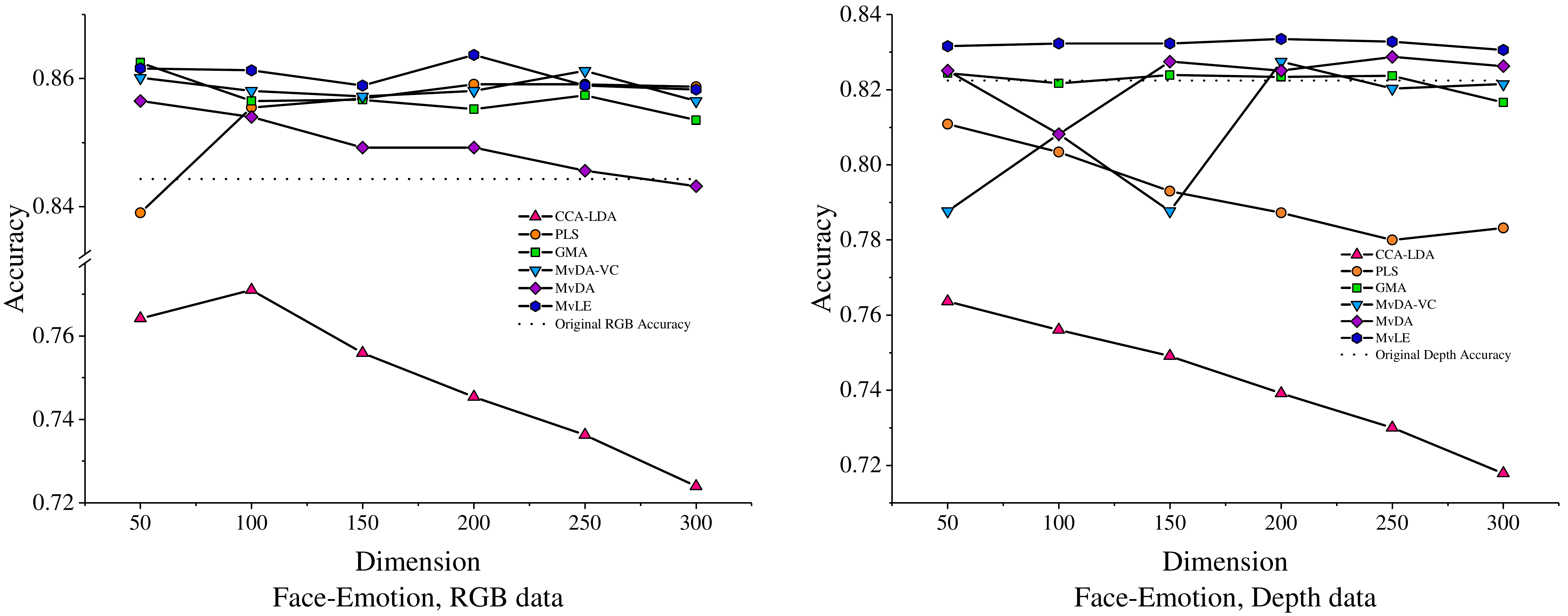}
\end{tabular}
\end{center}
\caption
{ \label{fig:face}
Change curve of comparison methods on different target dimension for the RGB-D face-emotion dataset. 
}
\end{figure*}

Firstly, 
we calculate the recognition accuracy of original RGB and depth features individually as the references for both datasets, 
as Table \ref{BefFusion} shows. 
Then we compare the proposed methods with several state-of-art methods, $i.e.$ PLS, GMA, MvDA-VC, MvDA and CCA-LDA, 
which use ELM as the classifier. 
All these methods use 2/3 data as training set, and the rest as test data. 
The experiments are randomly repeated for 5 times, 
and average accuracy are shown as Table \ref{VEcp} and Table \ref{FEcp}. 
Fig. \ref{fig:video}  and \ref{fig:face} 
show the change curve of different methods on our two datasets. 
For each figure, the left panel is on RGB data, the right panel is on the depth data. 
The dotted lines denote the recognition accuracy of RGB or depth data before multi-view learning. 

\begin{table*}[!htbp] 
  \begin{minipage}[b]{0.5\textwidth} 
    \centering 
\caption{Evaluation of Video-Emotion Dataset}
\begin{tabular}{p{1cm}<{\centering}p{0.6cm}<{\centering}p{0.6cm}<{\centering}p{0.6cm}<{\centering}p{0.6cm}<{\centering}p{0.6cm}<{\centering}p{0.6cm}<{\centering}p{0.7cm}<{\centering}}
\toprule
\makecell[c]{Methods} 
&{Feature} & 300 & 250 	  &	200    &  150   &  100   &   50   \\

\midrule
\multirow{2}*{\shortstack{CCA-\\LDA}} 
& RGB   & 0.1293 & 0.1485 & 0.1489 & 0.1491 & 0.1496 & 0.1493 \\ 
& depth & 0.1309 & 0.1477 & 0.1491 & 0.1486 & 0.1490 & 0.1486 \\ 

\midrule
\multirow{2}*{\shortstack{PLS}} 
& RGB   & 0.3716 & 0.3767 & 0.3746 & 0.3809 & 0.3697 & 0.3563 \\ 
& depth & 0.3201 & 0.3215 & 0.3180 & 0.3040 & 0.3129 & 0.3038 \\ 

\midrule
\multirow{2}*{\shortstack{GMA}} 
& RGB   & 0.1771 & 0.1742 & 0.1834 & 0.1590 & 0.1538 & 0.1617 \\ 
& depth & 0.1502 & 0.1464 & 0.1427 & 0.1389 & 0.1353 & 0.1542 \\ 

\midrule
\multirow{2}*{\shortstack{MvDA-\\VC}} 
& RGB   & 0.3812 & 0.3700 & 0.3636 & 0.3142 & 0.3333 & 0.2727 \\ 
& depth & \textbf{0.3262} & 0.3070 & 0.2927 & 0.2807 & 0.2624 & 0.2033 \\ 

\midrule
\multirow{2}*{\shortstack{MvDA}} 
& RGB   & 0.3427 & 0.3549 & 0.3166 & 0.3325 & 0.3038 & 0.2384 \\ 
& depth & 0.3111 & 0.3086 & 0.2927 & 0.2656 & 0.2376 & 0.2081 \\ 

\midrule
\multirow{2}*{\shortstack{MvLE}} 
& RGB   & \textbf{0.3917} & \textbf{0.3734} & \textbf{0.3892} & \textbf{0.4100} & \textbf{0.3949} & \textbf{0.3868} \\ 
& depth & 0.3260 & \textbf{0.3258} & \textbf{0.3322} & \textbf{0.3214} & \textbf{0.3244} & \textbf{0.3086} \\ 
\bottomrule
\end{tabular}
	\label{VEcp} 
  \end{minipage}%
  \begin{minipage}[b]{0.5\textwidth} 
    \centering
\caption{Evaluation of face-emotion dataset}
\begin{tabular}{p{1cm}<{\centering}p{0.6cm}<{\centering}p{0.6cm}<{\centering}p{0.6cm}<{\centering}p{0.6cm}<{\centering}p{0.6cm}<{\centering}p{0.6cm}<{\centering}p{0.7cm}<{\centering}}
\toprule
\makecell[c]{Methods} 
&{Feature} & 300 & 250 	  &	200    &  150   &  100   &   50   \\

\midrule
\multirow{2}*{\shortstack{CCA-\\LDA}} 
& RGB   & 0.7240 & 0.7363 & 0.7454 & 0.7559 & 0.7710 & 0.7642 \\ 
& depth & 0.7179 & 0.7300 & 0.7392 & 0.7491 & 0.7561 & 0.7636 \\ 

\midrule
\multirow{2}*{\shortstack{PLS}} 
& RGB   & \textbf{0.8587} & 0.8591 & 0.8591 & 0.8569 & 0.8555 & 0.8390 \\ 
& depth & 0.7832 & 0.7800 & 0.7872 & 0.7930 & 0.8034 & 0.8108 \\ 

\midrule
\multirow{2}*{\shortstack{GMA}} 
& RGB   & 0.8535 & 0.8574 & 0.8552 & 0.8567 & 0.8565 & \textbf{0.8625} \\ 
& depth & 0.8166 & 0.8237 & 0.8234 & 0.8239 & 0.8217 & 0.8244 \\ 

\midrule
\multirow{2}*{\shortstack{MvDA-\\VC}} 
& RGB   & 0.8565 & \textbf{0.8612} & 0.8581 & 0.8572 & 0.8581 & 0.8601 \\ 
& depth & 0.8215 & 0.8203 & 0.8275 & 0.7877 & 0.8082 & 0.7877 \\ 

\midrule
\multirow{2}*{\shortstack{MvDA}} 
& RGB   & 0.8432 & 0.8456 & 0.8492 & 0.8492 & 0.8540 & 0.8565 \\ 
& depth & 0.8263 & 0.8287 & 0.8251 & 0.8275 & 0.8082 & 0.8251 \\ 

\midrule
\multirow{2}*{\shortstack{MvLE}} 
& RGB   & 0.8583 & 0.8589 & \textbf{0.8637} & \textbf{0.8589} & \textbf{0.8613} & 0.8616 \\ 
& depth & \textbf{0.8306} & \textbf{0.8328} & \textbf{0.8335} & \textbf{0.8323} & \textbf{0.8323} & \textbf{0.8316} \\ 
\bottomrule
\end{tabular}
    \label{FEcp} 
  \end{minipage} 
\end{table*}

\begin{table}[!htbp]
\centering
\caption{Average Accuracy of Each Feature Before Fusion}
\begin{tabular}{p{2cm}<{\centering}p{1cm}<{\centering}p{1cm}<{\centering}}
\toprule
Datasets & RGB & depth \\
\midrule
video-emotion & 0.2983 & 0.2547 \\
face-emotion & 0.8443 & 0.8224 \\
\bottomrule
\end{tabular}
\label{BefFusion}
\end{table}

\subsection{Discussions}
Experimental results indicate that for both datasets, 
MvLE could not only performs better but also more stably as dimension decreases. 
For video-emotion dataset that has great nonlinearity, LDA based methods CCA-LDA and GMA perform poorly. 
Their recognition accuracy after multi-view learning is much lower than before. 
MvDA-VC and MvDA perform better with a higher target dimension, 
but when the target dimension decreases, the accuracy decreases quickly. 
This can be ascribed to their ignorance of inter-view and inner-view discriminant information. 
PLS tries to correlate the latent score of original space, as well as minimizing the variations of views in the common subspace. 
As a result, PLS gets a better and more stable performance, but PLS is difficult to extend to multi-view learning. 
By building global weighted graph and introducing the category discriminant information, 
the nonlinear method MvLE performs not only better, but also more stably when target dimension decreases. 
In case of 50 dimensions, the improvement of proposed method over MvDA-VC and MvDA is as much as 10.05 percent. 
But for face-emotion dataset that has poor nonlinearity, MvLE just has a weak advantage over other methods. 

\section{Conclusion} \label{conclusion}

In this paper, we propose a new nonlinear supervised multi-view learning method named MvLE and its out-of-sample extension MHON 
to perform the nonlinear task, which is RGB-D human emotion recognition. 
MvLE can map the training set of RGB-D data to a common subspace, 
and MHON is used to get the low-dimensional representations of test data. 
The new distance metric method BON can not only overcome the difference between views, 
but also introduce the category discriminant information. 
Moreover, we introduced a video-emotion RGB-D nonlinear dataset and a face-emotion RGB-D linear dataset to evaluate the proposed method. 
The experiment results indicate the effectiveness of our method in nonlinear data. 
In the future, we can apply MvLE and MHON to other machine learning tasks. 



\section*{Acknowledgment}

This study was funded by National Natural Science Foundation of People's Republic of China (No. 61672130, 61602082, 91648205, 31871106), 
the National Key Scientific Instrument and Equipment Development Project (No. 61627808), 
the Development of Science and Technology of Guangdong Province Special Fund Project Grants (No. 2016B090910001). 
All the authors declare that they have no conflict of interest.

Thanks for Shaohua Chen and Bin Zhan in the contribution of data acquisition of video-emotion dataset. 

\ifCLASSOPTIONcaptionsoff
  \newpage
\fi



\bibliographystyle{unsrt}
\bibliography{bibfile}

\vspace{-15 mm}
\begin{IEEEbiography}[{\includegraphics[width=1in,height=1.25in,clip,keepaspectratio]{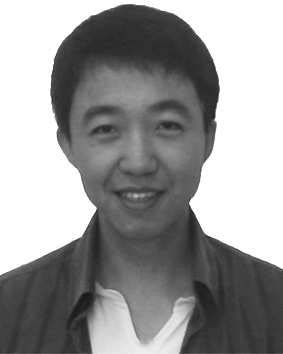}}]
{Shenglan, Liu}
received the Ph.D. degree in the School of Computer Science and Technology, Dalian University of Technology, China, in 2015. Currently, he is an associate professor with the School of Innovation and Entrepreneurship, 
Dalian University of Technology, China. 
His research interests include manifold learning, human perception computing. 
Dr. Liu is currently the editorial board member of Neurocomputing. 
\end{IEEEbiography}
\vspace{-10 mm}

\begin{IEEEbiography}[{\includegraphics[width=1in,height=1.25in,clip,keepaspectratio]{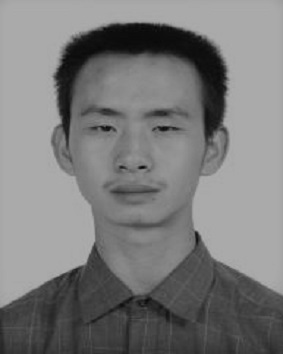}}]
{Shuai, Guo}
received the B.S. degree in the School of Computer Science and Technology from Dalian University of Technology, in 2017. 
Currently, he is working toward the M.S. degree in the School of Computer Science and Technology, 
Dalian University of Technology. 
His research interests include multi-view learning, dimensionality reduction, image and video learning. 
\end{IEEEbiography}
\vspace{-10 mm}

\begin{IEEEbiography}[{\includegraphics[width=1in,height=1.25in,clip,keepaspectratio]{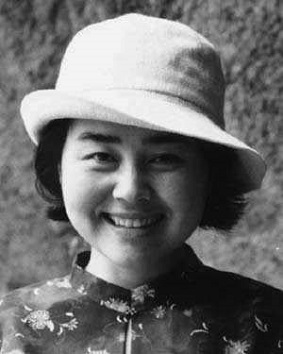}}]
{Hong, Qiao}
(SM’06) received the B.Eng. degree in hydraulics and control and the M.Eng. degree 
in robotics and automation from Xi'an Jiaotong University, Xi’an, China, 
and the Ph.D. degree in robotics control from De Montfort University, Leicester, U.K., in 1995. 
She was an Assistant Professor with the City University of Hong Kong, Hong Kong, 
and a Lecturer with the University of Manchester, Manchester, U.K., from 1997 to 2004. 
She is currently a Professor with the State Key Laboratory of Management 
and Control for Complex Systems, Institute of Automation, Chinese Academy 
of Sciences, Beijing, China. Her current research interests include robotics, 
machine learning, and pattern recognition.
\end{IEEEbiography}
\vspace{-10 mm}

\begin{IEEEbiography}[{\includegraphics[width=1in,height=1.25in,clip,keepaspectratio]{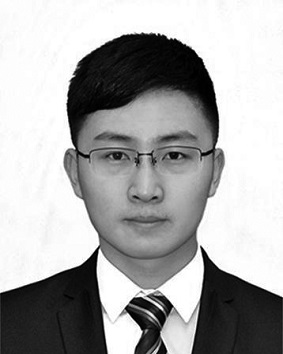}}]
{Yang, Wang}
is a Master degree candidate in Dalian University of Technology. 
His research interests include information retrieval, computer vision and machine learning. 
\end{IEEEbiography}
\vspace{-10 mm}

\begin{IEEEbiography}[{\includegraphics[width=1in,height=1.25in,clip,keepaspectratio]{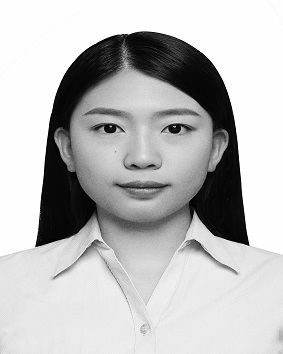}}]
{Bin, Wang}
is working toward the M.S. degree in the School of Computer Science and Technology, Dalian University of Technology. 
Her research interests include human action recognition and video retrieval. 
\end{IEEEbiography}
\vspace{-10 mm}

\begin{IEEEbiography}[{\includegraphics[width=1in,height=1.25in,clip,keepaspectratio]{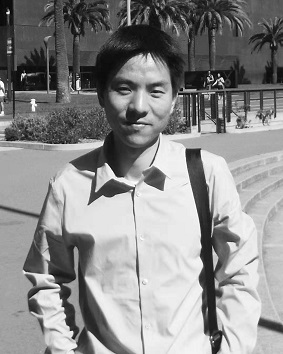}}]
{Wenbo, Luo}
received the Ph.D. degree in the School of Psychology, Southwest University, in 2009. 
Currently, he is a professor in the Research Center for Brain and Cognitive Neuroscience, Liaoning Normal University. 
His research interests include emotion and social neuroscience.
\end{IEEEbiography}
\vspace{-10 mm}

\begin{IEEEbiography}[{\includegraphics[width=1in,height=1.25in,clip,keepaspectratio]{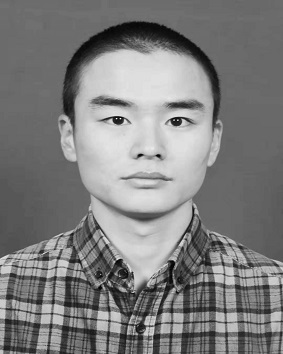}}]
{Mingming, Zhang}
received the M.S. degree in the Department of Psychology, Minnan Normal University, in 2017. 
Currently, he is working toward the Ph.D. degree in the Research Center for Brain and Cognitive Neuroscience, Liaoning Normal University. 
His research interests include emotion and social neuroscience.
\end{IEEEbiography}
\vspace{-10 mm}

\begin{IEEEbiography}[{\includegraphics[width=1in,height=1.25in,clip,keepaspectratio]{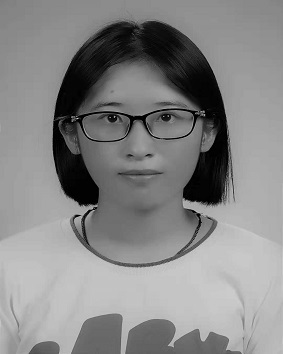}}]
{Keye, Zhang}
received the B.S. degree in the school of psychology, NanJing XiaoZhuang University, in 2016. 
Currently, she is working toward the M.Ed. degree in Research center for Brain and Cognitive Neuroscience, Liaoning Normal University. 
Her research interests include body recognition and social neuroscience.
\end{IEEEbiography}
\vspace{-10 mm}

\begin{IEEEbiography}[{\includegraphics[width=1in,height=1.25in,clip,keepaspectratio]{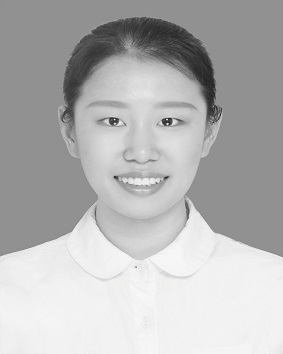}}]
{Bixuan, Du}
received the B.S. degree in the School of Psychology, Jiangxi Normal University, in 2017. 
Currently, she is working toward the M.S. degree in the Research Center for Brain and Cognitive Neuroscience, Liaoning Normal University. 
Her research interests include body recognition and social neuroscience.
\end{IEEEbiography}

\end{document}